\documentclass{INTERSPEECH2023}

\usepackage{caption}
\usepackage{wrapfig}
\usepackage{setspace}
\usepackage{cleveref}
\usepackage{enumitem}

\usepackage{multirow}
\usepackage{multicol}
\usepackage{xcolor}
\usepackage{esvect}
\usepackage{cite}
\usepackage{hyperref}

\makeatletter
\renewcommand{\section}{\@startsection
  {section}%
  {1}%
  {}%
  {-0.5\baselineskip}%
  {0.2\baselineskip}%
  {}}%

\renewcommand{\subsection}{\@startsection
  {subsection}%
  {2}%
  {}%
  {-0.1\baselineskip}%
  {0.1\baselineskip}%
  {}}%

\renewcommand{\subsubsection}{\@startsection
  {subsubsection}%
  {3}%
  {}%
  {-0.2\baselineskip}%
  {0.2\baselineskip}%
  {}}%

\g@addto@macro\normalsize{%
  \setlength\abovedisplayskip{5pt plus 2pt minus 2pt}
  \setlength\belowdisplayskip{5pt plus 2pt minus 2pt}
  \setlength\abovedisplayshortskip{4pt plus 2pt minus 2pt}
  \setlength\belowdisplayshortskip{4pt plus 2pt minus 2pt}
}

\makeatother

\Crefname{equation}{Eq.}{Eqs.}
\Crefname{section}{Sec.}{Sec.}

\interspeechcameraready

\title{RASR2: The RWTH ASR Toolkit for \\Generic Sequence-to-sequence Speech Recognition}
\name{Wei Zhou$^{1,2}$, Eugen Beck$^{2}$, Simon Berger$^{1,2}$, Ralf Schl\"uter$^{1,2}$, Hermann Ney$^{1,2}$}
\address{
  $^1$Machine Learning and Human Language Technology, Computer Science Department,\\
  RWTH Aachen University, 52074 Aachen, Germany \\
  $^2$AppTek GmbH, 52062 Aachen, Germany
}
\email{\{zhou, sberger, schlueter, ney\}@cs.rwth-aachen.de, ebeck@apptek.com}

\begin{document}

\maketitle

\begin{abstract}
Modern public ASR tools usually provide rich support for training various sequence-to-sequence (S2S) models, but rather simple support for decoding open-vocabulary scenarios only.
%while the decoding support is rather simple and only for open-vocabulary scenarios.
For closed-vocabulary scenarios, public tools supporting lexical-constrained decoding are usually only for classical ASR, or do not support all S2S models.
% To cover this gap
To eliminate this restriction on research possibilities such as modeling unit choice, we present RASR2 in this work, a research-oriented generic S2S decoder implemented in C++.
It offers a strong flexibility/compatibility for various S2S models, language models, label units/topologies and neural network architectures.
It provides efficient decoding for both open- and closed-vocabulary scenarios based on a generalized search framework with rich support for different search modes and settings.
%It also supports a flexible finite state automaton generation for forced alignment or training with external tools.
We evaluate RASR2 with a wide range of experiments on both switchboard and Librispeech corpora.
Our source code is public online.
\end{abstract}

\section{Introduction \& related work}
Sequence-to-sequence (S2S) modeling has become the major trend of automatic speech recognition (ASR).
Popular S2S modeling approaches include connectionist temporal classification (CTC) \cite{graves2016ctc}, recurrent neural network transducer (RNN-T) \cite{graves2012rnnt}, attention-based encoder-decoder (AED) models \cite{bahdanau2016end, chan2016listen} and possible variants thereof.
S2S models usually produce a very concentrated probability distribution, which requires smaller decoding effort in contrast to conventional ASR systems.
For a potentially open-vocabulary recognition, subword units are often adopted for both S2S models and the optional external language model (LM).
Without lexical constraint, this is usually called end-to-end ASR, which further simplifies search.

As a result, modern public ASR tools such as ESPNet \cite{watanabe2018espnet}, SpeechBrain \cite{speechbrain} and RETURNN \cite{zeyer2018returnn} usually put major effort on supporting various training techniques and NN architectures, while only providing decoding support for open-vocabulary scenarios, which usually adopts a simple global beam search with fixed beam size pruning.
Such tendency poses a strong restriction on research possibilities such as modeling unit choice, e.g. both phoneme/subword-based S2S models and word-based RNN/transformer \cite{transformer} LM can be individually trained with these tools, but they can not be jointly decoded within the framework.
\cite{hori17multiLM} proposed the multi-level LM decoding to incorporate a word-based LM into a subword-based end-to-end system by on-the-fly re-scoring upon emitting an end-of-word symbol.
As this is less straightforward for phonemes, \cite{wang20multiLMextend} extended it by introducing a lexical prefix tree.
The phoneme/subword LM score is then replaced by the word LM score upon tree exits, although LM-lookahead \cite{steinbiss94LMLA} can actually be applied here for more consistent pruning w.r.t. decision making.
More importantly, this decoder is not public, which also applies to \cite{variani2020hat} and possibly many other existing tools.
For closed-vocabulary scenarios, public ASR tools supporting lexical-constrained decoding are either only for classical ASR systems such as Kaldi\cite{kaldi} and previous RASR \cite{rybach2011:rasr, wiesler14rasr}, or only for CTC models such as flashlight \cite{kahn2022flashlight} (formerly wav2letter++ \cite{wav2letter++}) and K2 (next generation of Kaldi).

With this motivation, we present RASR2 in this work, a research-oriented generic S2S decoder implemented in C++.
RASR2 offers a strong flexibility/compatibility for various modeling unit choice/combination, modeling approaches/variants and NN architectures.
It implements a generalized search framework covering most existing search strategies for various S2S models.
With a rich support for different search modes and settings, it provides efficient decoding for both open- and closed-vocabulary scenarios. %sophisticated search methods
In contrast to weighted finite-state transducer (WFST)-based decoders, RASR2 performs lexical prefix tree search, which applies dynamic composition on the fly.
This makes it suitable for higher-order S2S models and/or LMs in one-pass decoding setups.
It also supports a flexible finite state automaton (FSA) generation for forced alignment or training with external tools.
We evaluate RASR2 with a wide range of experiments on both switchboard (SWB) \cite{swb} and Librispeech (LBS) \cite{libsp} corpora.
Our source code is public online\footnote{https://github.com/rwth-i6/rasr/tree/generic-seq2seq-decoder}
   
% verified over many papers crossing different topics

\vspace{-1mm}
\begin{figure}[h]
\begin{minipage}[b]{1.0\linewidth}
  \centering
  \centerline{\includegraphics[width=7cm]{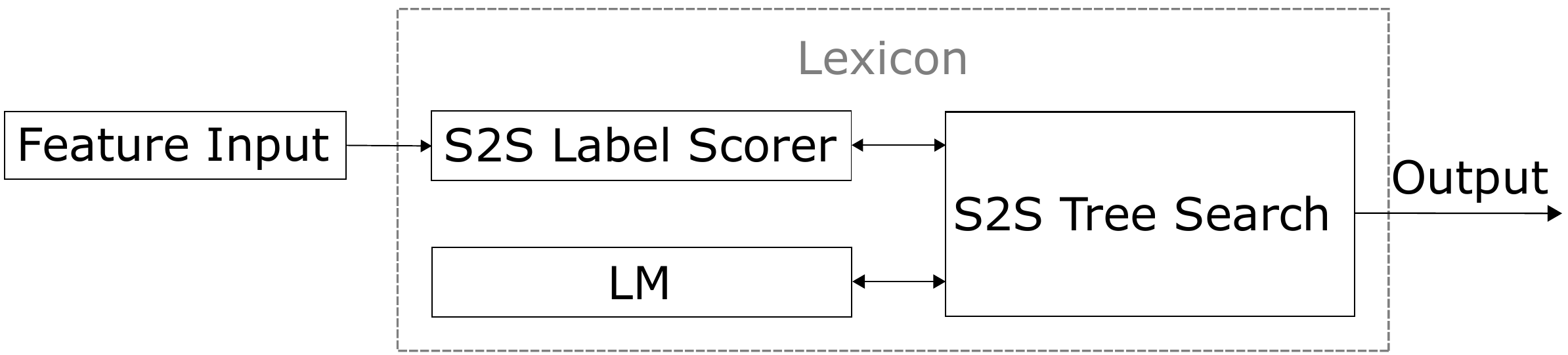}}
\end{minipage}
\caption{\it High-level structure of the S2S decoder}
\label{fig:system}
\vspace{-4.5mm}
\end{figure}

\section{Generic sequence-to-sequence decoder}
As an extension, RASR2 maintains most functionalities from previous versions \cite{rybach2011:rasr, wiesler14rasr}.
The major breakthrough is the newly introduced generic S2S decoder with high flexibility and a wide-range coverage of different models.
As shown in \Cref{fig:system}, its high-level structure contains the feature input, S2S label scorer and S2S tree search modules as well as an optional LM module.
% TODO refer to models
% any S2S model
For the feature input module, RASR2 provides a rich signal processing library that can be flexibly configured for various feature extraction, ranging from raw samples to MFCC, Log-Mel and gammatone \cite{schluter2007gt}, etc.
In the following, we describe the other three modules in more detail.

\subsection{Flexible lexical prefix tree}
RASR2 performs lexical prefix tree search, where the static search tree is fully defined by a broad-sense lexicon. 
% flat automaton,  lexicon: mostly a vocab file
The fundamental element in the lexicon is called {\it lemma}, whose components are shown in \Cref{tab:lemma}.
The static search tree is then constructed based on all {\it lemmata} in the lexicon.
For simplicity, we denote $\vv{a}$ as a sequence of S2S model labels $a$.
With each tree node holding one $a$, each $\vv{a}$ in a {\it lemma} generates a path through the tree with prefix sharing for compactness.
At the end of this path, a tree exit is then created to hold the transcription output and LM token (sequence).
Multiple $\vv{a}$ of one {\it lemma} are just path variants in the tree, and multiple exits on the same tree node are allowed.
Effectively, by varying label units on different levels, RASR2 supports flexible combination of models using same or different modeling units.
As a simplest case, the transcription output, S2S model and LM can all be defined on the same subword set $V$.
In this case, the search tree has just one root followed by $|V|$ leaf nodes, each with a single tree exit. 
This corresponds to the open-vocabulary scenario, which may also be referred to as lexicon-free in some literature.
On the contrary, a word-based {\it lemma} construction leads to the closed-vocabulary case.

% on-the-fly dynamic composition: larger LM

\subsection{Generalized search framework}
Based on the above search tree, RASR2 can perform search for various S2S models in a generalized framework.
At each decoding step $u$, a new hypothesis can be fully characterized as the tuple $(y_u, t_u, H(y_1^{u-1}), \textit{scores}, \textit{treeNode})$, where
\begin{itemize}[itemsep=-0.3mm]
\item $y_u$: a symbol predicted by the S2S model, including special ones such as blank $\epsilon$.
\item $t_u$: the position in the S2S model's encoder output $h_1^T$.
\item $H(\cdot)$ $=$ $(H_{\text{S2S}}, H_{\text{LM}})$: history functions of the S2S model and LM, both of which are model specific and can be empty. They may be used for model scoring as well as recombination.
\end{itemize}
For simplicity, we omit the traceback consideration here and assume the final hypothesis can always uniquely define the output transcription given the label topology and lexicon definition.
Existing search strategies for various modeling approaches mainly fall into three categories:
\begin{enumerate}[itemsep=-0.3mm]
\item time-synchronous search: all hypotheses are at the same global time position $t=u$. Typical models include CTC \cite{graves2016ctc}, monotonic RNN-T \cite{tripathi2019monoRNNT} and posterior HMM \cite{zhou2021phonemeTransducer, raissizhou2022:fullsum}.
\item label-synchronous search: all hypotheses are at the same output label position $u$, and $t_u$ may be omitted. Typical models include attention \cite{bahdanau2016end, chan2016listen} and segmental \cite{beck2018segmentalLVCSR, zhou2021seg-eq-trans} models.
\item alignment-synchronous search \cite{saon2020ALSD}: all hypotheses are at the same alignment position $u$, which is mainly for RNN-T \cite{graves2012rnnt}. Here $t_u = t_{u-1} + 1$ for $y_u=\epsilon$, otherwise $t_u = t_{u-1}$.
\end{enumerate}
These search strategies are actually all synchronized on the S2S model scoring level, i.e. decoding step $u$. 
They only differ at hypothesis expansion, while the rest operations in each decoding step are mostly the same.
Based on this, RASR2 unifies them into a common search framework as the following.

\subsubsection{Hypothesis expansion}
Within-tree hypothesis expansion is defined by the lexical prefix tree and label topology jointly, where the former defines reachable successors and the latter defines allowed transitions.
RASR2 supports flexible label topology configuration by fundamental topology operations such as loop, blank and vertical transitions.
Mutual exclusive cases such as vertical + loop and mutual inclusive cases such as vertical + blank are automatically checked.
This flexibility enables the study of different modeling variants, e.g. neural transducer with HMM topology \cite{zhou2021phonemeTransducer}.
Although it is natural to configure the topology according to the S2S model training, sometimes it may also be needed to decode with a certain variant, e.g. decode a CTC model without label loop, which can be useful for forced alignment as well.
Additional flags such as enabling $t_u$ as a latent position variable and forcing minimum loop occurrence \cite{raissizhou2022:fullsum} are also supported.

% merge paragraph ?
For hypotheses exiting a tree, they enter the next tree based on history $H_{\text{LM}}$.
Thus, $H_{\text{LM}}$ only needs to be handled on the search tree level instead of within-tree hypotheses.
Depending on the settings, hypotheses reaching the same node in the same tree may still have different $H_{\text{S2S}}$. 
For convenience, in the following, we refer to within-tree and exiting-tree hypotheses as `label' and `word-end' hypotheses, respectively.

\begin{table}[t]
\caption{\it Components of lexicon fundamental element: lemma}
%\vspace{-0.5mm}
\label{tab:lemma}
\setlength{\tabcolsep}{0.13em}
%\centering
\scalebox{0.9}{\parbox{1\linewidth}{%
\begin{tabular}{|l|c|}
\hline
{\it lemma} Components & Typical Unit  \\ \hline
\hspace{0.5mm} 1 \hspace{5mm} transcription output & word, subword  \\ \hline
\hspace{0.5mm} $\ge 1$ \hspace{1.5mm} sequences of S2S model labels: $\vv{a}$ & phonemes, subword(s) \\ \hline
\hspace{0.5mm} $0$ or $1$ LM token (sequence) & word, subword (sequence) \\ 
\hline
\end{tabular}}}
\vspace{-2mm}
\end{table}

\subsubsection{Scoring \& pruning}
All label hypotheses update scores from the S2S label scorer (\Cref{sec:labelScorer}). 
If LM is used, an optional LM-lookahead \cite{steinbiss94LMLA} score from the best reachable exits can also be added for pruning.
All word-end hypotheses then update scores from the scoring LM.
Note that RASR2 also supports different LMs for lookahead and word-end scoring, e.g. $n$-gram LM + full-context neural LM (NLM) for efficiency.

Three different kinds of pruning methods are mainly used:
\begin{itemize}[itemsep=-0.3mm]
\item score pruning: score threshold w.r.t. best hypothesis in beam
\item histogram pruning \cite{steinbiss94LMLA}: beam size upper limit
\item fixed beam size pruning: top-$k$-score hypotheses in beam
\end{itemize}
which are incorporated into two different kinds of beam search:
\begin{enumerate}[itemsep=-0.3mm]
\item simple global beam search: all levels of hypotheses, i.e. label, word-end and ended (\Cref{sec:ending}) hypotheses, are pruned together in one global beam. Either fixed beam size pruning or score + histogram pruning can be applied here.
\item hyp-level-individual beam search: each level of hypotheses has an individual beam for pruning. Only score + histogram pruning is applied here, which however can be configured individually for each beam.
\end{enumerate}
The global and individual beam search correspond to those simple and sophisticated decoders in modern and classical ASR tools, respectively.
This enables a strong compatibility as well as comparison with other tools.
Although both beam search variants work for either open- or closed-vocabulary scenarios, one might be favored over the other for specific cases, e.g. we find the hyp-level-individual beam search to be more efficient for the closed-vocabulary case.
For the open-vocabulary case, additional optimization is done to avoid redundant pruning for leaf nodes and exits.
We also support some additional tricks such as local score pruning within the same posterior output of the S2S model, max length pruning w.r.t. $T$, and special symbol threshold such as blank or end-of-sentence (EOS) \cite{hannun19eos}.

% TODO
% - check all covered 
% - CPU: very large search space (investigation)
% * GPU: also possible

\subsubsection{Recombination}
Similar as \cite{saon2020ALSD}, we also find it more efficient to apply recombination after pruning. 
RASR2 supports both maximization (Viterbi) and summation (full-sum) based recombination, which only differ at one operation in the dynamic programming recursion.
% change after double blind, our previous work
In \cite{zhou2020fullsum}, we showed that they have the same decoding efficiency.
Label hypotheses in the same tree (same $H_{\text{LM}}$) are recombined upon the same $(H_{\text{S2S}}, t_u)$.
Word-end hypotheses are recombined upon the same $(H_{\text{S2S}}$, $H_{\text{LM}}, t_u)$.
For Viterbi recombination , we support configurable history limit up to the context size of the models.
For full-sum recombination, at least one full context history from $H_{\text{S2S}}$ and $H_{\text{LM}}$ is required.

% history extension
% consistent with modeling(training)

\subsubsection{End processing}
\vspace{-0.1mm}
\label{sec:ending}
Another difference among time-/label-/alignment-synchronous search is ending hypothesis processing, where hypotheses can finish asynchronously at different $u$.
In the search framework of RASR2, this is predefined by each specific type of S2S model, such as $t_u=T$ for RNN-T and segmental models, and $y_u=$ EOS for attention models.
Ended hypotheses are stored separately without further expansion.
They are pruned against further ongoing hypotheses in the global beam search mode, or optionally pruned against each other in the individual beam search mode to control output size.
By default, no end processing is applied and the complete search stops at $u=T$.
If end processing is needed, the complete search stops when all live hypotheses are ended.
For deterministic scoring, RASR2 also performs early stopping if all ongoing hypotheses can not be better than the best ended one.

Asynchronous-ending may lead to length bias \cite{lengthBias2016} for unreliable score comparison, which can be eased by decoding heuristics such as length normalization.
RASR2 additionally supports a search-derived length modeling for more robust decision making \cite{zhou2020search}.

\subsubsection{Outputs and logs}
\vspace{-0.1mm}
For recognition output, RASR2 supports various formats including CTM, N-best and lattice, which can be further used for re-scoring or sequence discriminative training \cite{zhou2022transducerTrain}.
In addition, the decoding log file contains rich statistics of search as well as some timing information for analysis.

\subsection{S2S label scorer}
\label{sec:labelScorer}
The S2S label scorer module is an interface between search and the underlying S2S model.
It mainly manages model scoring, $H_{\text{S2S}}$ and some model type specifics.
For model scoring, it is designed to work with any encoder-decoder or encoder-only models without knowing the underlying NN details.
This is achieved by pre-compiling the NN into a computation graph with predefined I/O and 
operation collections, which are executed in order by RASR2.
Besides a strong compatibility with various S2S models and NN structures, this also allows flexibly modifying the S2S model graph for inference, such as adding an internal LM (ILM) \cite{variani2020hat, zhou2022transducerILM} and replacing some deterministic computation with embedding lookup.

\subsubsection{Scorer types}
\vspace{-0.1mm}
There are two major types of S2S label scorer:
\begin{enumerate}[itemsep=-0.3mm]
\item pre-computed scorer: scores are computed independent of search. This can be used to load externally computed scores, or scoring encoder-only models as well as some 1st-order encoder-decoder models with a small $|V|$. For the last case, the model graph needs to be modified to feed all $|V|$ contexts at each frame and output the $|V|^2$ scores.
\item encoder-decoder scorer: scores are computed on the fly with context feedback from hypotheses in search.
This is the base scorer for encoder-decoder models to control the model graph computation logics.
Specific types of S2S models just inherit from it with further specifications. This allows an easy extension of derived scorers to study various modeling variants.
\end{enumerate}
For the pre-computed scorer, RASR2 only needs to know the I/O placeholders of the model graph.
For the encoder-decoder scorer, the model graph should contain the following collections, where the decoder computation is single step based:
%the decoder computation in graph needs to be single step based,
%With determined I/O, this further eliminates differences among different S2S models including topology.
%and the graph should contain the following collections: 
\begin{itemize}[itemsep=-0.5mm]
\item {\it encode{\_}ops}: encoder forwarding with output kept in graph
\item {\it decoder{\_}input{\_}vars}: mostly label context to be fed in
\item {\it update{\_}ops}: optional update after feeding decoder input
\item {\it decode{\_}ops}: decoder forwarding to compute scores
\item {\it decoder{\_}output{\_}vars}: mostly scores to be fetched
\item {\it state{\_}vars}: optional decoder hidden states, i.e. stateful vs. stateless (mostly limited context). If existing, they are fed in together with label context and fetched together with scores. They are hold as pointers by $H_{\text{S2S}}$ together with label context. 
% all details hidden to RASR2
\end{itemize}
The {\it encode{\_}ops} is only called once for each input sequence (full/chunk-wise), while the decoder computation may be executed multiple times at each step $u$.
This depends on the total number of hypotheses in search and the configurable maximum number of hypotheses to feed in for a single run of decoder computation.
For $H_{\text{S2S}}$, it is also configurable to include loop and/or blank into the dependency.
 
% history managing
% e.g. transducer variants, HMM topology 
 
% TODO: combined scorer  

\subsubsection{Backends}
\vspace{-0.1mm}
The model graph may use different backends to perform computation on GPU and/or CPU.
Currently we mainly support Tensorflow \cite{tensorflow} backend, and are extending to support ONNX\footnote{https://github.com/onnx/onnx} backend for a wider coverage of models trained with different tools.
Note that search is always performed on CPU to allow investigation of very large search space, e.g. \cite{zhou2020search}

\subsection{LM}
RASR2 also supports various LM types for both one-pass decoding and lattice/N-best re-scoring:
\begin{itemize}[itemsep=-0.3mm]
\item n-gram LM in arpabo format
\item NLM such as RNN and transformer LM: this has a similar design of model graph and backends as \Cref{sec:labelScorer}. 
Speed-up tricks from our previous work \cite{beck2020transformerlm, eugen19lstmlm1pass} are also included.
% one advantage against WFST-based decoder : on-the-fly dynamic composition: thus larger LM
\item combined LM: score combination of several LMs
\end{itemize}
Note that RASR2 even supports to configure separate LMs for scoring, lookahead and recombination.
To study some special cases of recombination, we also provide some simple simulations such as 0-gram LM and full-context LM.

\vspace{-0.4mm}  
\section{Flexible Finite State Automaton}
Another useful feature of RASR2 is the flexible generation of a finite state automaton (FSA) for a given sentence.
The flexibility comes at two folds.
Firstly, the broad-sense lexicon allows a flexible definition of $\vv{a}$ for each {\it lemma} including label units as well as path variants (e.g. pronunciation/segmentation) for the same word.
Secondly, RASR2 allows to specify various label topologies for the FSA, including HMM, CTC and recurrent neural aligner (RNA) \cite{sak2017rna}.
The resulting FSA can be used to perform forced alignment with S2S models.
It can also be exported as a simple list of edges ({\it from, to, label, weight}) to external training tools to perform from-scratch sequence-level cross-entropy (CE) training using the forward-backward algorithm. 
% + RETURNN, cite our papers ?
Different aspects can then be studied, such as topology comparison \cite{raissizhou2022:fullsum} and acoustic-based subword learning \cite{zhou2021ADSM}.

\vspace{-0.4mm}  
\section{Experiments}
To reflect the rich functionality, compatibility and flexibility of RASR2, we conduct a wide range of experiments across different S2S models, NN architectures, label units, LMs and search settings on both switchboard (SWB) \cite{swb} and Librispeech (LBS) \cite{libsp} corpora.

\subsection{Open-vocabulary efficiency comparison}
Modern ASR tools usually provide a simple decoder with global beam search and fixed beam size pruning, which commonly works well for subword-based S2S models in open-vocabulary scenarios.
Here we verify if RASR2 is also competitive under this setting.
We use RETURNN \cite{zeyer2018returnn} as a representative for comparison, which uses the same backend as RASR2 and performs purely GPU-based decoding.
This is done on the SWB corpus using a standalone byte-pair encoding \cite{sennrich16BPE} (BPE)-based attention model with BLSTM encoder + LSTM decoder \cite{zeineldeen2021ilm}.
We disable the multi-sequence recognition for RETURNN.
Both word error rate (WER) and real time factor (RTF) results are shown in \Cref{tab:swb-attention}, where RASR2 shows competitive efficiency.

% maintain fast for standalone E2E system using simple beam search

\begin{table}[t]
\caption{\it Efficiency comparison for decoding a standalone BPE-based attention model on Hub5'00; RTF on {\footnotesize GTX-1080Ti GPU}}
\vspace{-0.5mm}
\label{tab:swb-attention}
\setlength{\tabcolsep}{0.67em}
%\centering
\scalebox{0.8}{\parbox{1\linewidth}{%
\begin{tabular}{|c|c|c|c|c|}
\hline
\multirow{2}{*}{Tool} & \multirow{2}{*}{Search \& Pruning} & \multicolumn{2}{c|}{Hub5'00 {\scriptsize WER[\%]}} & RTF \\
  & & {\hspace{1mm} SWB } & CH & (GPU) \\ \hline
RETURNN \cite{zeyer2018returnn} & \multirow{2}{*}{\hspace{-3mm}\shortstack[c]{global beam search\\+ fixed beam size 12}} & \multirow{3}{*}{8.8} & \multirow{3}{*}{17.4} & 0.042\\ \cline{1-1} \cline{5-5}
\multirow{2}{*}{RASR2} & & & & 0.042\\ \cline{2-2} \cline{5-5}
                                & \hspace{4mm} + local score pruning & & & 0.039 \\
\hline
\end{tabular}}}
\end{table}

\begin{table}[t]
\caption{\it Subword S2S models + closed-vocabulary constraint {    }}
\vspace{-0.5mm}
\label{tab:LBS+word}
\setlength{\tabcolsep}{0.72em} 
%\centering
\scalebox{0.8}{\parbox{1\linewidth}{%
\begin{tabular}{|c|c|c|c|c|c|c|}
\hline
S2S   & \multicolumn{2}{c|}{Word} & \multicolumn{2}{c|}{LBS dev {\scriptsize WER[\%]}} & \multicolumn{2}{c|}{LBS test {\scriptsize WER[\%]}} \\
Model & Lexicon  &      LM        & {\hspace{1mm} clean } & other  & {\hspace{1mm} clean } & other \\ \hline
\multirow{3}{*}{\shortstack[c]{Subword\\CTC}} & \multicolumn{2}{c|}{no}      & 4.0 & 11.1 & 4.2 & 11.3 \\ \cline{2-7}
                     & \multirow{2}{*}{yes} &  no   & 3.5 & \phantom{1}9.7  & 3.9 & \phantom{1}9.9 \\ \cline{3-7}
                     &                   &   4-gram & 2.7 & \phantom{1}6.8  & 3.7 & \phantom{1}7.3 \\ \hline
\multirow{4}{*}{\shortstack[c]{Subword\\Transducer}} & \multicolumn{2}{c|}{no} & 2.7 & \phantom{1}6.9 & 2.9 & \phantom{1}7.1 \\ \cline{2-7}
 & \multirow{3}{*}{yes} &  no     & 2.7 & \phantom{1}6.9 & 3.1 & \phantom{1}7.1 \\ \cline{3-7}
 &                    &   4-gram  & 2.4 & \phantom{1}5.6 & 2.7 & \phantom{1}6.1 \\ \cline{3-7}
 &                      & Trafo   & 1.8 & \phantom{1}4.0 & 2.1 & \phantom{1}4.4 \\
\hline
\end{tabular}}}
\vspace{-0.5mm}
\end{table} 

\begin{figure}[t!]
\begin{minipage}[b]{1.0\linewidth}
  \centering
  \centerline{\includegraphics[width=7cm]{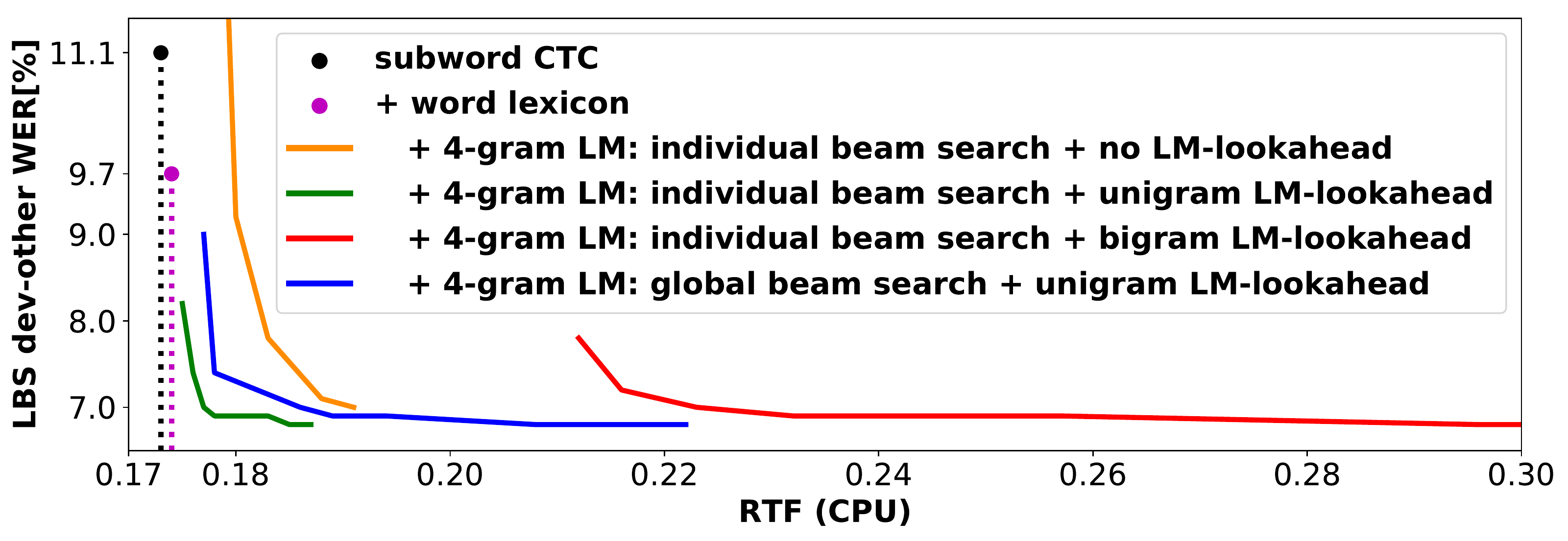}}
\end{minipage}
\vspace{-4.5mm}
\caption{\it WER vs. RTF (on {\footnotesize Intel Xeon CPU E5-2630 v4 @2.20GHz})}
\label{fig:wer_rtf}
\vspace{-4mm}
\end{figure}

\subsection{Closed-vocabulary constraint}
\label{sec:closeVocab}
Here we investigate the effect of applying closed-vocabulary constraint to subword-based S2S models for the LBS task.
We use 5k acoustic data-driven subword modeling (ADSM) \cite{zhou2021ADSM} units.
We construct word-based {\it lemmata} for the lexicon according to the official LBS vocabulary.
Segmentation variants for the same word from the ADSM output are also included into the lexicon.
We evaluate two scenarios, i.e. word lexicon only and word lexicon+LM, for both subword-based CTC and monotonic RNN-T models. 
Both models use a $12 \times 512$ conformer\cite{Gulati20conformer} encoder.
The transducer model adopts a $1 \times 1024$ LSTM decoder, and applies ILM subtraction \cite{variani2020hat} when external LM is used.
The results are shown in \Cref{tab:LBS+word}.

For the CTC model, the word lexicon alone already largely improves over the open-vocabulary baseline, while an additional 4-gram word LM further boosts the performance dramatically.
This is as expected due to the lack of context modeling of the CTC model.
For the transducer model, there is no benefit from the word lexicon alone.
% hyps restriction: corrected some spellings, but also mistaked some OOVs
Although the additional 4-gram word LM gives some improvement, we suspect that the gain will be smaller with an increasing amount of transcribed data as well as possible ILM adaptation \cite{rnntTextAdapt2021, meng22ILMA} on text data. 
Further results with a word-based transformer LM also show worse performance than the open-vocabulary evaluation with a subword-based transformer LM (last row of \Cref{tab:recomb}).

With a super fast inference speed, CTC models are often adopted for production systems.
Adding closed-vocabulary constraint to a subword CTC does not impose pronunciation modeling problem as for phonemes, but rather a specification of hypothesis space.
Besides WER improvement with additional context modeling, this also brings additional flexibility for easy context biasing, which can be useful for many applications.
Therefore, we further study how to maintain the above improvement w/o cost of speed.
We mainly investigate LM-lookahead (no/unigram/bigram) and score-based beam search (hyp-level-individual/global beam search) settings, both of which become more important with word LM applied.
Their WER vs. RTF (CPU) plots are shown in \Cref{fig:wer_rtf}, where the individual beam search + unigram LM-lookahead shows the best efficiency, closely approaching the speed of a standalone CTC.

% LA , search : also verify on phoneme transducer
% many deployed transducer only use limited context size
% - anyway some more powerful LM is needed/used, but NLM slow 
% - lexicon + 4gram word LM: one comfortable solution for better WER + flexibility 
%    ngram word LM still bring additional info/modeling power + flexibility w/o loss of speed

\begin{table}[t]
\caption{\it Effect of Viterbi vs. full-sum recombination in decoding; at least one full-context model from the S2S model and LM}
\vspace{-0.5mm}
\label{tab:recomb}
\setlength{\tabcolsep}{0.25em}
%\centering
\scalebox{0.8}{\parbox{1\linewidth}{%
\begin{tabular}{|c|c|c|c|c|c|c|c|}
\hline
S2S   & \multirow{2}{*}{LM} & Decoding & \multicolumn{2}{c|}{LBS dev {\scriptsize WER[\%]}} & \multicolumn{2}{c|}{LBS test {\scriptsize WER[\%]}} \\
Model & &  Recombination & {\hspace{1mm} clean } & other & {\hspace{1mm} clean } & other \\ \hline
\multirow{2}{*}{\shortstack[c]{Phoneme\\Transducer}} & \multirow{2}{*}{\shortstack[c]{Word\\Transformer}} & Viterbi & 1.7 & 3.5 & 2.0 & 4.0 \\ \cline{3-7}
& & full-sum & 1.7 & 3.5 & 1.9 & 4.0 \\ \hline
\multirow{4}{*}{\shortstack[c]{Subword\\Transducer}} & \multirow{2}{*}{no} & Viterbi & 2.7 & 6.9 & 2.9 & 7.1 \\ \cline{3-7}
  & & full-sum & 2.7 & 6.9 & 2.9 & 7.1 \\ \cline{2-7}
& \multirow{2}{*}{\shortstack[c]{Subword\\Transformer}} & Viterbi & 1.8 & 3.9 & 1.9 & 4.3 \\ \cline{3-7}
  & & full-sum & 1.8 & 3.9 & 1.9 & 4.3 \\
\hline
\end{tabular}}}
\end{table}

\subsection{Recombination: Viterbi vs. full-sum}
\label{sec:recomb}
For full-context models, recombination based on limited histories becomes inappropriate.
This motivates revisiting full-sum instead of Viterbi recombination, which is also more consistent with some model training such as transducer models.
In \cite{zhou2020fullsum}, we showed that full-sum recombination gives consistently small improvement over the Viterbi counterpart for hybrid HMM systems with LSTM LM.
Here we also re-visit this for 3 cases: 
\begin{enumerate}[itemsep=-0.8mm]
\item full-context $H_{\text{LM}}$ only: 1st-order phoneme-based conformer transducer \cite{zhou2022transducerTrain} + word-based transformer LM \cite{irie19trafolm}
\item full-context $H_{\text{S2S}}$ only: subword-based transducer w/o LM
\item full-context $H_{\text{S2S}}$ and $H_{\text{LM}}$: subword-based transducer + subword-based transformer LM (ILM subtraction applied)
\end{enumerate}
We apply the recipe in \cite{zhou2022transducerTrain} to train the transducer models. The subword transducer is the same as in \Cref{sec:closeVocab}. The results on LBS are shown in \Cref{tab:recomb}, where Viterbi recombination gives the same performance as full-sum in almost all cases.

\begin{table}[t]
\caption{\it WER[\%] of phoneme-based conformer transducer + word-based transformer (Trafo) LM; 1-pass vs. re-scoring}
\vspace{-0.5mm}
\label{tab:nlm}
\setlength{\tabcolsep}{0.27em}
%\centering
\scalebox{0.8}{\parbox{1\linewidth}{%
\begin{tabular}{|c|c||c|c|c|c||c|c|c|}
\hline
\multirow{2}{*}{LM} & \multirow{2}{*}{Decoding} & \multicolumn{2}{c|}{LBS dev} & \multicolumn{2}{c||}{LBS test} & \multirow{2}{*}{\footnotesize Hub5'00} & \multirow{2}{*}{\footnotesize Hub5'01} & \multirow{2}{*}{\footnotesize RT'03}\\
  &              &  clean & other  & clean & other &  &  & \\ \hline
4-gram & 1-pass (lattice) & 2.4 & 5.5 & 2.8 & 6.0 & 9.9 & 10.1 & 11.4 \\ \cline{1-1}
\multirow{2}{*}{Trafo} & + re-scoring & 1.8 & 3.7 & 2.0 & 4.2 & 9.0 & \phantom{1}9.3 & 10.2 \\ \cline{2-9}
  & 1-pass & \bf 1.7 & \bf 3.5 & \bf 1.9 & \bf 4.0 & \bf 8.9 & \phantom{1}\bf 9.1 & \bf 10.0 \\
\hline
\end{tabular}}}
\vspace{-3mm}
\end{table}

\subsection{Full-context NLM decoding: one-pass vs. re-scoring}
Full-context NLM one-pass decoding, especially for closed-vocabulary scenarios, is not widely supported with all public tools.
Instead, using n-gram LM one-pass decoding to generate lattice/N-best for re-scoring is often applied.
Here we use RASR2 to compare these two approaches.
%re-visit NLM one-pass decoding vs. lattice re-scoring.
We use the same models as in case 1 of \Cref{sec:recomb} (full-sum recombination).
The results on both LBS and SWB corpora are shown in \Cref{tab:nlm}, where one-pass decoding consistently outperforms re-scoring.

%NLM 1pass not always supported
%- especially for WFST-based decoder, only re-scoring
%- closed-vocabulary 

\vspace{-1mm}
\section{Conclusions}
\vspace{-0.3mm}
In this work, we presented RASR2, a research-oriented generic sequence-to-sequence decoder in C++ for ASR.
We gave details about its design principle including a generalized search framework, 
rich support for various search modes and settings,
and strong flexibility/compatibility for different models, NNs and label units/topologies.
These aspects, plus efficiency, are further verified with a wide range of experiments on both SWB and LBS corpora.
Our source code is public online.

%
%future work
%- 1pass multi-label level model combination: aiming at best performance
%- generic segmental model
%- multi utterances
%- streaming

\vspace{-0.55mm}
\begin{center}
\bf Acknowledgements
\end{center}
\vspace{-0.35mm}
%\footnotesize
\scriptsize
This work was partially supported by a Google Focused Award and by NeuroSys which, as part of the initiative ``Clusters4Future'', is funded by the Federal Ministry of Education and Research BMBF (03ZU1106DA).
The work reflects only the authors' views and none of the funding parties is responsible for any use that may be made of the information it contains.
We also thank Albert Zeyer for useful discussions.

\bibliographystyle{IEEEtran}
\bibliography{refs}

\end{document}